\pdfoutput=1

\documentclass[11pt]{article}

\usepackage[]{acl}

\usepackage{times}
\usepackage{latexsym}
\usepackage{booktabs}
\usepackage[T1]{fontenc}

\usepackage[utf8]{inputenc}

\usepackage{inconsolata}
\usepackage{caption}
\usepackage{graphicx}
\usepackage{amsmath, amssymb}
\usepackage{booktabs}

\title{Cross-lingual Data Augmentation for Document-grounded Dialog Systems in Low Resource Languages}

\author{
Qi Gou, Zehua Xia, Wenzhe Du\\
  State Key Laboratory for Novel Software Technology, Nanjing University, China \\
  \{qi.gou, zehuaxia, gowott\}@smail.nju.edu.cn
  }

\newcommand\Model{CLEM}

\begin{document}
\maketitle
\begin{abstract}

This paper proposes a framework to address the issue of data scarcity in Document-Grounded Dialogue Systems(DGDS). Our model leverages high-resource languages to enhance the capability of dialogue generation in low-resource languages. Specifically, We present a novel pipeline {\Model} (Cross-Lingual Enhanced Model) including adversarial training retrieval (Retriever and Re-ranker), and Fid (fusion-in-decoder) generator. To further leverage high-resource language, we also propose an innovative architecture to conduct alignment across different languages with translated training. Extensive experiment results demonstrate the effectiveness of our model and we achieved 4th place in the DialDoc 2023 Competition. Therefore, CLEM can serve as a solution to resource scarcity in DGDS and provide useful guidance for multi-lingual alignment tasks.

\end{abstract}

\section{Introduction}



Document-Grounded Dialogue System (DGDS) is a meaningful yet challenging task, which not only allows content accessible to end users via various conversational interfaces, but also requires generating faithful responses according to knowledge resources. 

However, in real-world scenarios, we may not have abundant resources to construct an effective dialogue system due to the low resources of some minority languages such as Vietnamese and French. Previous works only consider building a DGDS in high-resource languages with rich document resources such as English and Chinese \cite{feng2021multidoc2dial,fu2022doc2bot}, which is contrary to real-world situations. Extensive minority languages struggle to build well-founded chatbots due to the low resource of documents.

Therefore, how to generate evidential responses under a scarce resources setting deserves our attention. To address this issue, we propose a novel architecture to leverage high-resource languages to supplement low-resource languages, in turn, build a fact-based dialogue system. Thus, our model can not only handle high-resource scenarios but also generate faithful responses under low-resource settings.Our key contributions can be split into three parts: 
\begin{itemize}
    \item We proposed a novel framework, dubbed as {\Model}, including adversarial training Retriever, Re-ranker and Fid (fusion-in-decoder) generator.
    \item We presented the novel architecture of translated training and three-stage training.
    \item Extensive results demonstrated the effectiveness of {\Model}. Our team won the 4th place in the Third DialDoc Shared-task competition.
\end{itemize}

\section{Related Work}
\paragraph{Document Grounded Dialogue System} is an advanced dialogue system that requires the ability to search relevant external knowledge sources in order to generate coherent and informative responses.
 To evaluate and benchmark the performance of such systems, existing DGDS datasets can be broadly classified into three categories based on their objectives: 1) \textbf{Chitchat}, such as WoW \cite{wow}, Holl-E \cite{holle}, and CMU-DoG \cite{cmu_dog_emnlp18}. These datasets typically involve casual and open-ended conversations on various topics; 2) \textbf{Conversational Reading Comprehension (CRC)}, which requires the agent to answer questions based on understanding of a given text passage. Examples of CRC datasets include CoQA \cite{coqa}, Abg-CoQA \cite{guo2021abgcoqa}, and ShARC \cite{sharc}; and 3) \textbf{Information-seeking Scenarios}, such as Doc2dial \cite{doc2dial}, Multidoc2dial \cite{feng2021multidoc2dial}, and Doc2bot \cite{fu2022doc2bot}, where the agent needs to retrieve relevant information from one or more documents to address a user's query. 

\paragraph{Cross-lingual Data Augmentation} 
 has emerged as an effective approach to address the challenges of multilingual NLP tasks \cite{rw1, rw2, rw3, rw4, rw5}. Particularly in low-resource language settings, DA has demonstrated its usefulness \cite{rw6, rw7, rw8}. Explicit DA techniques mainly involve translation-based templates, such as word-level adversarial learning \cite{rw9} and designed translation templates \cite{rw6, rw7}. Implicit data augmentation techniques, on the other hand, focus on modeling instead of expanding datasets like representation alignment \cite{rw10}, knowledge distillation \cite{rw11} and transfer learning \cite{rw12}.

\section{Task Description}
\label{task description}
\paragraph{Formulation.} We aim to improve the performance of DGDS in low-resource languages (Vietnamese and French). Formally, given labeled set $D=\{x_i,p_i,r_i\}, i\in [1, N_{D}]$ , where $N_D$ denotes the number of data and $x_i,p_i,r_i$ denotes the input, grounding passage and response. Note that the input is obtained by concatenating the current turn and previous context. In addition, we have access to some high-resource language labeled datasets $U$ with size $N_U$, where $N_U \gg N_D$. Our goal is to explore how to utilize high-resource datasets to enhance performance in low-resource languages (Vietnamese and French). 

We have access to two large datasets, namely Multidoc2dial \cite{feng2021multidoc2dial} for English and Doc2bot for Chinese\cite{fu2022doc2bot}. To fully take advantage of these high-resource datasets to enhance the performance in French and Vietnamese, we conducted translated training and generated pseudo-labeled training sets in Vietnamese and French. Specifically, we utilized the Baidu API\footnote{\href{https://fanyi-api.baidu.com/api/trans/product/index}{https://fanyi-api.baidu.com/api/trans/product/index}} 
and Tencent API\footnote{\href{https://www.tencentcloud.com/products/tmt}{https://www.tencentcloud.com/products/tmt}} to translate English and Chinese into French and Vietnamese, separately. Notably, English and French are Indo-European languages, indicating a common ancestral language, and Chinese and Vietnamese share historical and cultural connections and have influenced each other. Our methodology involved augmenting the training set by translating 5000 English examples into French and 5000 Chinese examples into Vietnamese. After filtering out instances of poor quality and excessive length, we ultimately derived 4980 En-Fr and 4908 Zh-Vi pseudo examples.

Now we have three training data, cross-lingual training data $D$, translated pseudo data $D^{'}$ and downstream fine-tuning data $D^{t}$. We will show how to use these data in Section \ref{sec: trainig}. 
And the statistics are presented in Table \ref{tab: Statistics}.

\begin{table}[t]
    \centering
    \begin{tabular}{l| c}
         \toprule
         data  & number of turns \\ 
         \midrule
         Chinese corpus & 5760 \\
         English corpus & 26506 \\
         Shared-Task/train & 3446 (Vi) and 3510(Fr)\\
         Zh-Vi &  4908 \\
         En-Fr &  4980 \\
         Shared-Task/dev & 95(Vi) and 99(Fr) \\
         Shared-Task/test & 94(Vi) and 100(Fr) \\
         \bottomrule
    \end{tabular}
    \caption{Statistics of provided datasets. Chinese and English corpus is provided by the third workshop committee of DialDoc. Zh-Vi and En-Fr means the number of translated data from Chinese to Vietnamese and from English to French respectively.}
    \label{tab: Statistics}
\end{table}

\section{Methodology}
 We adopt the Retrieve-Rerank-Generation architecture \cite{glass-etal-2022-re2g,zhang2023coarse} and incorporate adversarial training into both the Retriever and Re-ranker components. To address the low-resource DGDS scenario, we propose a novel three-stage training approach.
 
\subsection{Passage-Retriever With FGM}
Given an input $x$, the retriever aims to retrieve the most relevant top-k documents $\{z_i\}^k_{i}$ from a large candidate pool. 
We follow the schema of conventional Dense Passage Retrieval (DPR) \cite{karpukhin2020dense} for passage retrieval: 
\begin{align*}
  s(q) &= \operatorname{XLM-R}_{1}(q)\\
  s(z) &= \operatorname{XLM-R}_{2}(z) \\  
  p_{\phi}(z|q) & \propto \operatorname{dot}[s(q)^{\top} s(z)] 
\end{align*}

To improve multi-lingual performance further, where the encoder is initialized from  XLM-RoBERTa \cite{conneau2019unsupervised} denote as $\operatorname{XLM-R}$ which are used to convert question templates into dense embedding vectors for passage retrieval. Sub-linear time search can be achieved with a Maximum Inner Product Search (MIPS) \cite{shrivastava2014asymmetric}. 

In addition, inspired by FGM \cite{FGM}, we extend the adversarial training to document retrieval. We apply infinitesimal perturbations on word embeddings to increase the learning difficulty by constructing adversarial examples. Based on this, the passage retriever is regularized and has better generalization performance since it has to retrieve the correct relevant documents under the attack of adversarial examples. 

\subsection{Passage-Reranker with FGM}
Given a shortlist of candidates, the goal of Reranker is to capture deeper interactions between a query $x$ and a candidate passage $p$. Specifically, the query $x$ and passage $p$ are concatenated to form the input for XLM-RoBERTa \cite{conneau2019unsupervised}. And the pooler output of XLM-RoBERTa is considered as similarity score:
\begin{align*}
P(p|q) = \operatorname{SoftMax}\left(\operatorname{Linear}\left(\operatorname{XLM-R}([p,q])\right)\right) 
\end{align*}
As in the previous stage, we still employed FGM \cite{FGM} to add perturbations to word embeddings.
\subsection{Knowledge-Enhancement Generation}
The generator aims to generate correct and factual responses according to the candidates of passages. The key problem is how to leverage the knowledge of passage candidates as much as possible. 
we adopt Fusion-in-Decoder(FiD) \cite{izacard2021leveraging} as our response
generator. During generation, FiD will first encodes every input with multiple passages independently through encoder, and then decodes all encoded feature jointly to generate final response. Concisely, the decoder has extra Cross Attention on more passages feature. This is significant because it is equivalent to improve grounding passage accuracy from top-k to top-n. Note that $k \ll n$ due to the CUDA memory limitation.

Since prompt-learning is effective in generation proved by previous work \cite{wei2021finetuned}, we also adopt this way by adding the prompt to the front of input query. We choose \textit{"please generate the response:" } as our prompt, so the final input of generator is ~"\texttt{prompt <query> query <passage> passage}", where \texttt{<prompt>} and \texttt{<passage>} are special tokens.

\subsection{Training Process}\label{sec: trainig}
Our training process consists of three stages. In the first stage, we use all available Chinese and English training corpora to pre-train the model, aiming to develop its primary cross-lingual perception capability. We incorporate downstream fine-tuning data in this stage as well. We denote this stage as $T(D+D^t)$, where $T$ represents training.

In the second stage, we train the model using translated pseudo data, which includes both noisy data and downstream fine-tuning data. We denote this stage as $T(D^{'} + D^t)$.

Finally, we fine-tune the model from the second stage on downstream low-resource training data. We denote this stage as $F(D^t)$, where $F$ represents fine-tuning.

Therefore, the complete training process can be represented as $T(D+D^t)T(D^{'} + D^t)F(D^t)$. In the Experiment section, we also explore other training processes, such as two-stage training and direct fine-tuning.


 \begin{table}[t]
    \centering
    \begin{tabular}{l|c}
    \toprule
         Model & Total \\ 
         \midrule
         Baseline & 156.42 \\
        {\Model}&  \textbf{201.0913} \\ 
    \bottomrule
    \end{tabular}
    \caption{Performance of \Model{} on Test set}
    \label{tab:main}
\end{table}

\begin{table*}[t]
    \centering
    \begin{tabular}{l|c|c|c|c}
         \toprule
         \Model & F1 & BLEU & ROUGE & Total \\
         \midrule
       { \Model}-Full & \textbf{66.51} & \textbf{57.45} & \textbf{64.38} & \textbf{188.34} \\ 
       { \Model}(two-stage) & 65.52 & 55.23 & 63.15 & 183.9 \\
       { \Model}(fine-tune) & 63.76 & 53.41 & 61.47 & 178.64 \\
       { \Model}(two-stage w/o Zh-Vi) &  64.24 & 54.51 & 62.18 & 180.93\\
       { \Model}(two-stage w/o En-Fr) &  61.99 & 51.21 & 60.28 & 173.48 \\
       { \Model}(w/o prompt) & 64.34 & 55.12 & 62.31 & 181.77 \\
        \bottomrule
    \end{tabular}
    \caption{Ablation results of {Model}on Development set. Here, the best are marked with \textbf{Bold}. Two-stage means we do not use original Chinese and English data. Fine-tune means we just use downstream training data.}
    \label{tab:abaltion}
\end{table*}

\begin{table}[tbp]
    \centering
    \begin{tabular}{l|c|c|c|c}
        \toprule
         \Model & R@1 & R@5 & R@20 & MRR@5 \\ 
        \midrule
         retrieval & 0.57 & 0.78 & 0.87 & 0.65 \\
         retrieval\dag & 0.62 & 0.77 & 0.87 & 0.68 \\
         re-rank & 0.74 & 0.84 & 0.87 & 0.78 \\
         re-rank \dag & \textbf{0.76} & \textbf{0.85} & 0.87 & \textbf{0.79}\\
         \bottomrule
    \end{tabular}
    \caption{Effect of FGM on Development set, where \dag means we use adversarial training}
    \label{tab:fgm}
\end{table}

\section{Experiments and Results}\label{experiment}
In this section, we will introduce our datasets and baseline system. Additionally, we will demonstrate the effectiveness of each component in our methodology, such as adversarial training and the novel training process.
\subsection{Datasets}

We train {\Model} on the given shared task datasets, containing Vietnamese (3,446 turns), 816 dialogues in French (3,510 turns) and a corpus of 17272 paragraphs in ModelScope\footnote{\url{https://modelscope.cn/}}, where each dialogue turn is grounded in a paragraph from the corpus. Moreover, we also utilize Chinese (5760 turns) and English (26,506 turns) as additional training data.
\subsection{Baseline System}
The baseline follows the pipeline of Retrieval, Re-rank and Generation. It simply uses DPR \cite{karpukhin2020dense} as retriever and Transformer Encoder \cite{Trasformer} with a linear layer as re-ranker.



\subsection{Result and Analysis}

We evaluate the generation results based on token level F1, SacreBLEU and Rouge-L. The final result is the sum of them. As shown in Table \ref{tab:main}, {\Model} has a significant improvement by 28\% on total result compared to strong baseline, which demonstrates the effectiveness of our method. 

\subsubsection{Ablation Study}
We study the impact of different components of{\Model}, where the results are given in Table \ref{tab:abaltion}.
\paragraph{Training process}
we compare {\Model} with two-stage training and fine-tuning directly. The former only contains translated corpus without original Chinese and English data, which can be denoted by $T(D^{'}+D^t)F(D^t)$. While the latter means we only use downstream fine-tuning data denoted by $F(D^t)$. 
From the first three lines of Table \ref{tab:abaltion}, we can observe that {\Model} has superior performance than two-stage training which means {\Model} can leverage cross-lingual corpus to do a better language alignment for downstream training and get a better initialization. Not surprisingly, two-stage training outperforms fine-tuning directly which echos the Translated Training \cite{rw2} 
\paragraph{Different pseudo corpus}
    As described in section \ref{task description}, we leverage two translated pseudo corpus Zh-Vi and En-Fr. We also study the impact of each set with two-stage training. From 4th and 5th line of Table \ref{tab:abaltion}, the performance without Zh-Vi(Chinese to Vietnamese) and En-Fr(English to French) will decrease, which proved that the translated corpus is useful for shared task.
\paragraph{Without prompt}
    We also run the experiments without prompt to explore the impact of prompt. From the last line of Table \ref{tab:abaltion}, the performance of {\Model} will decrease sharply. 
\paragraph{Without FGM}
    We also explore the effectiveness of FGM \cite{FGM} at retriever and re-ranker. Results are listed in Table \ref{tab:fgm}. We can observe significant improvements from retrieval to re-rank which prove the effectiveness of re-rank.
\section{Conclusion}
This paper introduces \Model{}, a novel pipeline for document-grounded dialogue systems that uses a "retrieve, re-rank, and generate" approach. To address the issue of low performance due to limited training data, we extend the adversarial training to the document Retriever and Re-ranker components. Additionally, \Model{} leverages high-resource languages to improve low-resource languages and develops a new training process under data-scarce settings.

Experimental results demonstrate that \Model{} outperforms the strong, competitive baseline and achieved 4th place on the leaderboard of the third DialDoc competition. These findings provide a promising approach for generating grounded dialogues in multilingual settings with limited training data and further demonstrate the effectiveness of leveraging high-resource languages for low-resource language enhancement.

\section*{Acknowledgements}

\bibliography{anthology, main}
\bibliographystyle{acl_natbib}

\appendix

\section{Experiments Hyperparameters}
\label{sec:appendix}
\subsection{Hyper-parameters for retriever}
train\_batch\_size=128 \\ 
epochs=50 \\
max\_input\_length=512 \\
dropout=0.1 \\
weight\_decay=0.1 \\
warmup\_steps=1000 \\
gradient\_checkpoint\_segments=32 \\
optim=adam \\
learning\_rate=4e-05 \\
preKturns=2 \\
\subsection{Hyper-parameters for re-reanker}
learning\_rate=2e-5 \\
dropout=0.1 \\
epochs=20  \\
train\_batch\_size=1 \\
accumulation\_steps=32 \\
weight\_decay=0.1 \\
warmup\_steps=1000 \\
max\_input\_length=512 \\
passages=20 \\
preKturns=2 \\
\subsection{Hyper-parameters for generator}
learning\_rate=2e-4 \\
dropout=0.1 \\
epochs=20  \\
accumulation\_steps=16 \\
max\_grad\_norm=1 \\
train\_batch\_size=1 \\
accumulation\_steps=1 \\
weight\_decay=0.1 \\
warmup\_steps=1000 \\
max\_input\_length=1024 \\
max\_output\_length=128 \\
beam\_size=3 \\
passages4gen=5 \\
preKturns=2 \\
\end{document}